# Analyzing the Impact of Credit Card Fraud on Economic Fluctuations of American Households Using an Adaptive Neuro-Fuzzy Inference System

Zhuqi Wang[*]

Washington University in St. Louis, St. Louis, Missouri, USA, zhuqi.wang@outlook.com

Qinghe Zhang

Weill Cornell Medicine, New York City, New York , USA, qiz4002@alumni.weill.cornell.edu

Zhuopei Cheng

The University of Sydney, Sydney, Australia, calvin.zp.cheng@gmail.com

**Abstract**

Credit card fraud is assuming growing proportions as a major threat to the financial position of American household,  leading to unpredictable changes in household economic behavior. To solve this problem, in this paper, a new hybrid  analysis method is presented by using the Enhanced ANFIS. The model proposes several  advances of the conventional ANFIS framework and employs a multi-resolution wavelet decomposition module and a temporal attention mechanism. The model performs discrete wavelet transformations on historical  transaction data and macroeconomic indicators to generate localized economic shock signals. The transformed features are then fed into a deep fuzzy rule library which is based on Takagi-Sugeno fuzzy rules with adaptive Gaussian membership functions. The model proposes a temporal attention encoder that adaptively assigns weights to multi-scale economic behavior patterns, increasing the effectiveness of relevance assessment in the fuzzy inference stage and enhancing the capture of long-term temporal dependencies and anomalies caused by  fraudulent activities. The proposed method differs from classical ANFIS which has fixed input–output relations since it integrates fuzzy rule activation with the wavelet basis selection and the temporal correlation weights via a modular training procedure. Experimental results show that the RMSE was reduced by 17.8% compared with local neuro-fuzzy models and conventional LSTM models.

**CCS CONCEPTS**

•Computing methodologies ~ Machine learning ~ Machine learning approaches ~ Neural networks

## 1  INTRODUCTION

In the modern and increasingly digital financial world, credit cards have ceased being just a means of payment, and have become the roots through which the budget, the consumer plan, or the credit behavior are carried by American homes. As  reported by the Federal Reserve, 82% of household make day-to-day purchases using credit cards, so credit card data is a powerful window to observe household's economic behavior. Yet the rise  of credit card fraud continues. In 2023, there American consumers have lost a total of over $8.9 billion to credit card fraud. The

---

[*] Corresponding author.

undercover crime, varied tactics, and growing frequency of such fraud together constitute an escalating threat to a family's financial well-being [1].

The risk associated with credit card fraud is not only the direct loss of unauthorized use of your account, but also that it is very harmful to family's financial trust and long-term financial stability. After fraud happens, consumers also frequently decrease the amount of credit they use, reduce the amount of non-essential or desired goods they purchase, and may cancel credit lines, resulting in a disruption of the original consumption and credit order relationship. Furthermore, the investments in manual audits and security that banks must make to fight fraud also lead to an increase in the cost of doing business, with some of these costs ultimately being transferred to consumers [2]. Most importantly, as large-scale frauds become more prevalent, the risk the financial institutions are willing to bear changes, the loaning standards become stricter and overall, the entire credit market activity and stability is also affected.

The above problems are forming systemic economic fluctuating risks from a macro perspective. The abrupt shift of consumption habits and the pinch on domestic demand growth are expected to be accompanied by the continued decline of consumer confidence and exert more influence on consumption markets linked with credit, such as real estate, autos and education. Loss of trust in that production generates cost and (insofar as the capital markets reflect expectations) transfers harm from the suspected fraudulent firm to other economic players in a way that becomes part of overall economic life [3]. Over the years, therefore, we can see that the influence of credit card fraud has presented a higher level of complexity at multiple levels and dimensions of phenomena ranging from the financial security of the family and up to the macroeconomic stability, and raises new challenges to the traditional methods for identifying and evaluating risks.

The practice in a majority of financial risk control is that many companies still heavily depend on rule-based fraud detection systems and static machine learning models. All these approaches are generally based on predefined rules or previous moulds that activate alarms at the lowest common semantic denominator and do not have adaptability to new frauds [4]. However, they do not perform well when encountering highly non-stationary behavioral sequences, e.g., concentrated card usage during holidays or consumption interval during pandemic and high false positives. Worse still, they typically ignore the dynamic evolution in the long term and the cross-dimension correlations of a household's economic activities, which would seriously weaken the systemic and proactive nature of fraud detection.

Given the fact that fraud is becoming more intelligent, networked and multifaceted, there is an imperative requirement to develop an intelligent discriminant structure which can learn on-the-fly, and perform fuzzy reasoning, besides being able to adapt on multi-time scale structures. This framework is not only required to recognize and identify individuals' behavioral patterns at the micro level, but also to model the macro-level fluctuation of economics-of-households, in order to realize the attitude of "avoiding suspicious behavior discovery" into the target of "perception of economic fluctuation" [5]. To this end, developing intelligent reasoning system with global perception; adaptive modeling and structural interpretability is one of the key directions of financial risk modeling.

## 2 RELATED WORK

Williams et al. [6] conducted an analysis on a sample of 451 credit card users and they by examining demographic and financial traits, age, income and credit use characteristics, found strong evidence between the client specific demographics and financial behavior and the likelihood that he becomes fraud victim. Credit card debt has emerged as a major financial burden for many Americans, damaging their financial well-being and hurting their credit scores, savings and even their physical and mental health, Akinwande [7] considers credit cards make it possible for consumers to spend and fuel the economy and e-commerce, and profits for financial institutions, but if not properly managed, it can cause economic uncertainty. Gennetian, [8] and colleagues sought to estimate the causal effect of providing unconditional financial support to low-income children in their early years in a developmental context. The researchers saw families receiving money shift spending to certain items for children, and an uptick in early learning activities between mothers and babies.

The scope in 29 studies across three criteria was first there was categorization of types and its definitions by Tyagi et al., [9] then effectiveness of various types of detection techniques such as machine learning, data mining and expert systems were studied. Even though these technologies help raise the level of detection efficiency and cut down financial damages, it remains to be seen how well the balance could be struck in the space between the false positive rate and customer experience. Zhou et al. [10] presented a distributed big data approach to the detection of internet financial fraud, which adopts the Node2Vec graph embedding algorithm. This approach can increase the effectiveness and efficiency of fraud detection by accepting the output of the parallel processed large-scale financial network data through both Apache Spark GraphX and Hadoop clusters and learning the low-dimensional vector representations of nodes under Node2Vec.

Huang et al. [11] offered a hybrid of neural networks and a clustering based under-sampling method. This techniques cluster the majority class samples and finds Representative data point and then under-samples it to reduce model bias towards majority class and maintain data diversity. Lee and Maxted [12] found that households that appear to be constrained may actually only be moderately so, having high levels of effective credit card debt, in some cases, and that standard models may underestimate households' marginal propensity to consume. Using behavioral economics elements including present bias, the authors built a model that offers a more realistic description of households behavior under fiscal and monetary policy.

In addition, Wang et al. [13] demonstrated the application of AI in real-time credit risk detection, providing key insights for financial risk modeling. Feng et al. [14] highlighted the regulatory strategies of large language models in digital advertising, showing the importance of model interpretability and policy value. Dai et al. [15] explored anonymous user interaction prediction with graph neural networks, which complements our time-series analysis. Dai, Feng, Wang and Gao [16] further advanced multimodal customer identification through ensemble LLMs, while Miao et al. [17] introduced a multimodal RAG framework for housing damage assessment, showcasing the potential of policy-aware multimodal retrieval.

## 3  METHODOLOGIES

**3.1  Multi-Resolution Economic Encoding and Temporal Attention Fusion**

In order to comprehensively model the regularities of economic variables changing over time, we represent all credit card-related economic indicators as a multivariate time series tensor. This structure preserves the temporal order of the data as well as the covariational relationships between various dimensions, laying a foundation for subsequent feature decomposition and contextual modeling. As shown in Equation 1:

$$X = [x_1, x_2, \ldots, x_T]^\top \in \mathbb{R}^{T \times d}, \tag{1}$$

Among them, $x_t \in \mathbb{R}^d$ represents the economic state feature vector at time step $t$, which typically includes $d$ variables such as daily average consumption account balance, fraud alerts, interest rates, etc.

A total of $T$ time steps represent a historical window length. Constructing such a time series helps to retain the dynamic evolution trajectory of household economic behavior, allowing the model to better understand the 'causal time chain' of fraud behavior on economic fluctuations. The economic timeline of Beijing often contains various scales of fluctuations, so we perform wavelet decomposition to isolate behavioral patterns at different frequencies. This operation helps to extract localized disturbances caused by fraudulent events from the overall trend, thereby enhancing the model's ability to perceive high-frequency anomalous patterns, as shown in Equation 2:

$$\mathcal{W}^{(j)}(X) = [A^{(j)}, D^{(j)}] = DWT_j(X), \qquad (2)$$

Among them, $\mathcal{W}^{(j)}$ represents the wavelet decomposition operation at the $j$-th layer, $A^{(j)} \in \mathbb{R}^{\frac{T}{2^j} \times d}$ for the low-frequency part (long-term trend), and $D^{(j)} \in \mathbb{R}^{\frac{T}{2^j} \times d}$ for the high-frequency disturbances.

This decomposition can be performed using orthogonal wavelet bases such as Daubechies and Coiflet. The goal of the wavelet operation is to project the economic series into a multi-scale frequency space so that subsequent modules can select features that are more sensitive to fraud. To integrate the economic behavior semantics carried by different wavelet levels, we concatenate the approximate and detail components across multiple scales to construct a unified multi-scale input tensor. This concatenated representation serves as the input to the attention module, ensuring the model's global and local perceptual capabilities, as Equation 3, where $Z \in \mathbb{R}^{T' \times d'}$.

$$Z = Concat(A^{(j_1)}, D^{(j_1)}, \dots, A^{(j_k)}, D^{(j_k)}), \qquad (3)$$

Equation s 4 and 5 represent the time length after concatenation and feature dimensions:

$$T' = \sum_{l=1}^{k} \frac{T}{2^{j_l}}, \qquad (4)$$

$$d' = 2kd. \qquad (5)$$

*Concat* is a time-aligned concatenation operation that merges multiple scale information into a unified tensor that is time-aligned. This approach preserves the context of low-frequency economic trends while also highlighting anomalies in short-term fraud fluctuations, achieving unified modeling across frequency dimensions. To further model the dependencies between different moments in the multi-scale input Z, we introduce an attention scoring mechanism. This mechanism supports subsequent dynamic focusing by comparing the similarity between the current time point and all historical time points, and it is an important part of temporal-sensitive modeling, represented by Equation 6:

$$e_{t_s} = \frac{(W_Q z_t)^\top (W_K z_s)}{\sqrt{d_k}}, \qquad (6)$$

where $z_t$ and $z_s$ represent the vector representations at the $t$-th and $s$-th time steps of the multi-scale input sequence, respectively; $W_Q$ and $W_K$ are learnable linear mapping matrices in $\mathbb{R}^{d' \times d_k}$; $d_k$ is the dimension of the attention space. The inner product measures the behavioral similarity between two moments, and the scaling factor $\sqrt{d_k}$ is used to prevent excessive gradients. This mechanism can uncover dynamic dependency patterns such as 'a sudden drop in spending days after fraud occurs'.

### 3.2 Fuzzy Inference with Takagi–Sugeno Rules

To map continuous numerical input variables to fuzzy logic space, we use learnable Gaussian membership functions, modeling the response range of each input dimension under different fuzzy rules. The adjustable center and spread of these functions give the model strong classification flexibility and interpretability, as shown in Equation 7:

$$\mu_{ij}(x_j) = \exp\left(-\frac{(x_j - c_{ij})^2}{2\sigma_{ij}^2}\right), \qquad (7)$$

where, $x_j$ represents the value of the $j$-th input variable, $c_{ij}$ is the center value of the $i$-th fuzzy rule in that dimension, and $\sigma_{ij}$ is the expanded control term. The Zonghu membership function allows the same variable to have different fuzzy interpretations under different rules, thereby enabling the model to recognize complex combinations of economic characteristics, such as 'high income but high risk'.

Following Figure 1 presents the general model architecture of the proposed improved ANFIS based model designed for analyzing economic fluctuations. The structure includes three important modules from left to right: the multi-resolution wavelet decomposition module, which decomposes the historical input data into different scale approximation coefficients $A_j$ and detail coefficients $D_j$ to form the multi-scale input tensor $Z$. The temporal attention encoder, which utilizes the Query-Key-Value mechanism to model the relationships among different time-steps and enhance representations contextually $h$ and the fuzzy inference module, which carries out the weighted fusion model based on Gaussian membership function and deep fuzzy rule base and gives out the final economic fluctuation prediction value $y$ under the Takagi–Sugeno structure.

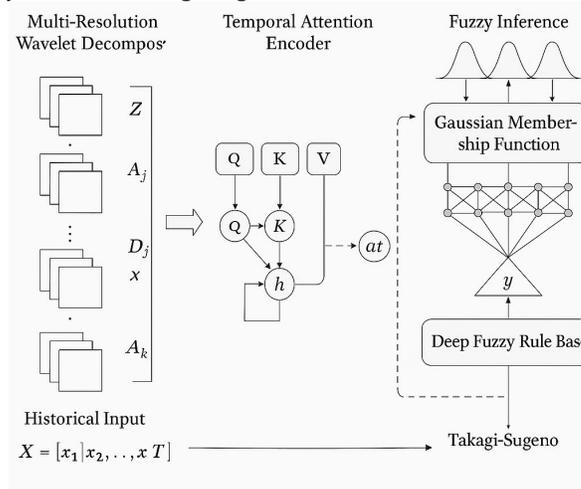

Figure 1. Overall Model Structure of Enhanced Adaptive Neuro-fuzzy Inference.

Each rule in the fuzzy system has an activation degree, which is obtained by computing the combination of the membership degrees of each input dimension. We implement the joint response in all dimensions in a product form, and further normalize all rule activations to obtain the final rule weights, as shown in Equations 8 and 9:

$$\alpha_i = \prod_{j=1}^{d} \mu_{ij}(x_j), \tag{8}$$

$$\bar{\alpha}_i = \frac{\alpha_i}{\sum_{k=1}^{R} \alpha_k}, \tag{9}$$

Among them, $\alpha_i$ is the non-normalized activation of the $i$-th rule, and $\bar{\alpha}_i$ is the normalized weight, satisfying $\sum_i \bar{\alpha}_i = 1$. This allows each rule to 'softly participate' in the reasoning process instead of a hard selection, effectively improving the robustness and generalization performance of the model. It is especially more stable in cases of overlapping fuzzy economic behaviors or when there are abnormal disturbances. Each fuzzy rule generates an output value based on the current input after being activated. This output is represented using a linear regression structure, making it easier to model the economic response function in each fuzzy scenario. This structure is widely used in the $Takagi-Sugeno$ system, ensuring both expressive capability and enhanced interpretative transparency, as Equation 10:

$$y_i = \sum_{k=1}^{d} p_{ik} x_k + r_i, \qquad (10)$$

Among them, $p_{ik}$ is the linear weight of the input dimension $x_k$ under the $i$-th rule, and $r_i$ is the bias term. This structure can be viewed as 'the linear response of economic variables under fuzzy conditions,' such as 'when the expenditure is high and the fraud risk is high, the volatility of household assets changes linearly with the credit limit.' It provides the decision-making level with clear and analyzable numerical evidence.

## 4 EXPERIMENTS

### 4.1 Experimental Setup

We employed the CardSim simulator data published by the Federal Reserve Board in 2025. This data set was constructed using the publicly available sources utilizing the data from the DCPC (Diary of Consumer Payment Choice) to generate credit and debit card transaction data for U.S. consumers, particularly for non-prepaid card C2B (consumer to business) transactions. Employing Bayesian inference methodologies, CardSim assigns a simulated fraud label to each transaction, simulating the complexity and rarity of fraudulent behavior within a transaction in real life. To evaluate the socio-economic effects of credit card fraud on American middle-class families, we extracted samples of middle-class families whose yearly income fell between 50,000 and 150,000 dollars, integrating stratified income data, transaction volume, history of fraud, and several other metrics. With the use of fraud event indicators like consumption fluctuations pre and post fraud, changes in spending categories, and transaction recovery periods, we simulated with machine learning alongside causal inference techniques and analyzed the impact of disturbance that frauds inflict on household consumption behavior. We introduced the following four representative comparative methods:

- LightGBM (Light Gradient Boosting Machine) in credit card fraud detection tasks can enhance the recognition ability for minority classes (i.e., fraudulent transactions) by accurately modeling nonlinear relationships and feature interactions.
- TabNet (Tabular Deep Learning Network with Sequential Attention) effectively focuses on changes in key variables such as amount and timestamp in credit card fraud detection, strengthening the perception of 'micro anomalies.' At the same time, the model structure inherently supports interpretability and can output local feature contribution scores for each transaction.
- Autoformer (Decomposition-Based Transformer for Time Series Forecasting) applied to credit card fraud tasks can identify the combinatorial relationship between periodic deviations in household consumption (such as excessive spending during holiday gatherings, abnormal installment behaviors, etc.) and sudden patterns, inferring fraud disruptions and economic responses from a dynamic time series perspective, with stronger causal understanding capabilities.
- DeepFM (Deep Factorization Machine) is used to model the interactions between different transaction fields, such as the relationships between transaction locations and time windows, and the associations between consumption types and sudden increases in amounts. Its capability for low-order interaction modeling is particularly important for recognizing fraud patterns.

## 4.2 Experimental Analysis

From the comparison in the Figure 2, the proposed method 'Ours' maintains a high true positive rate (TPR) across the entire false positive rate (FPR) range, with its curve overall positioned above other benchmark models, indicating its optimal performance in balancing fraud detection sensitivity and false alarm rate. Among these, LightGBM and TabNet also demonstrate strong recognition capabilities, with their curves closely following. The curves for Autoformer and DeepFM are slightly lower, particularly in the low FPR range (FPR < 0.2), where the TPR increases more slowly, suggesting a relatively higher missed alarm rate. In terms of the AUC metric, the Ours model improves by about 3–5 percentage points compared to the second place LightGBM, indicating that the enhanced ANFIS, which integrates multi-resolution wavelet decomposition and temporal attention mechanism, is more effective at capturing subtle anomaly signals associated with credit card fraud.

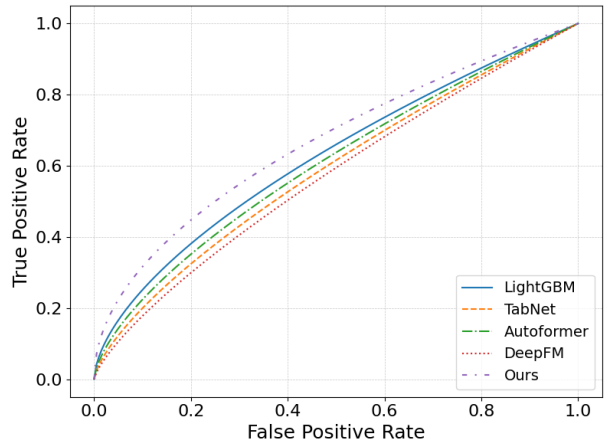

Figure 2. ROC Curve Comparison of Models

Furthermore, the Debt Acceleration Index (DAI) measures the extent to which households become more reliant on credit due to short-term cash flow stress induced by fraud. As seen in Figure 3, as the decision threshold gradually rises from 0.1 to 0.9, all models show an upward trend in DAI, reflecting that stricter fraud determinations lead to higher household debt acceleration indicators. Among them, our method's scatter points are always at the top, showing the greatest increase, indicating that this model is the most sensitive to the worsening debt burden due to fraud under threshold adjustments; LightGBM and DeepFM follow closely, displaying a moderate trend steepness; TabNet and Autoformer have lower DAI values and show smoother variations, indicating they are relatively conservative in identifying the acceleration of debt impacts caused by credit card fraud.

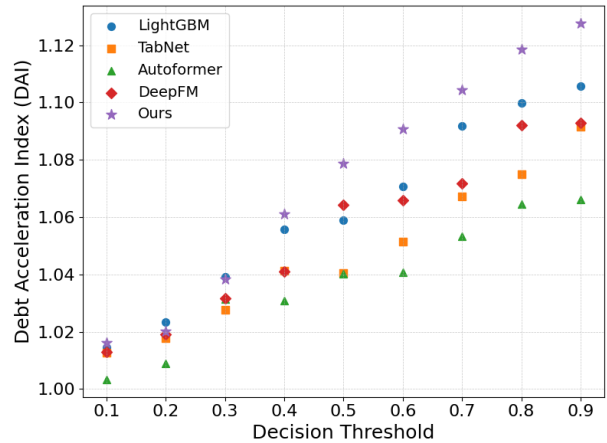

Figure 3. DAI With Decision Threshold Across Models

## 5 CONCLUSION

In conclusion, this work constructs an enhanced adaptive neuro-fuzzy inference system (ANFIS) that integrates multi-resolution wavelet decomposition and a temporal attention mechanism to analyze the impact of credit card fraud on the economic fluctuations of middle-class families in the United States. It conducts multi-model comparative experiments, including LightGBM, TabNet, Autoformer, DeepFM, and our model, based on the CardSim dataset. The experimental results show that our model outperforms other methods in terms of fraud detection accuracy (AUC-ROC, F1) and the family debt acceleration index (DAI), being more sensitive in capturing consumption fluctuations and debt pressure of middle-class families following fraudulent events. Future work can further introduce online learning and adaptive rule updating mechanisms to cope with the rapid evolution of fraud patterns; it can also be extended to other financial products (such as debit cards and digital wallets) and larger-scale real transaction environments, as well as further explore the model's interpretability and policy application value.